\documentclass{article}
\usepackage{icais_conference,times}
\iclrfinalcopy

\usepackage{amsmath,amsfonts,bm}

\def\eqref#1{equation~\ref{#1}}

\def\1{\bm{1}}

\DeclareMathAlphabet{\mathsfit}{\encodingdefault}{\sfdefault}{m}{sl}
\SetMathAlphabet{\mathsfit}{bold}{\encodingdefault}{\sfdefault}{bx}{n}

\usepackage{amsmath,amssymb}
\usepackage{booktabs}
\usepackage{array}
\usepackage{tabularx}
\usepackage{longtable}
\usepackage{graphicx}
\usepackage{float}
\usepackage{xcolor}
\usepackage{colortbl}
\usepackage{microtype}
\usepackage{url}
\usepackage{hyperref}

\definecolor{trizblue}{HTML}{0B4F8A}
\definecolor{triznavy}{HTML}{17324D}
\definecolor{trizgreen}{HTML}{2E9D65}
\definecolor{trizyellow}{HTML}{F4C95D}
\definecolor{trizmagenta}{HTML}{B65A8D}
\definecolor{trizorange}{HTML}{E8843A}
\definecolor{trizrow}{HTML}{F1F5F8}
\hypersetup{colorlinks=true,linkcolor=trizblue,citecolor=trizgreen,urlcolor=trizmagenta}

\newcommand{\triza}{TRIZ$^a$}
\newcommand{\complete}{\textcolor{trizgreen}{\textbf{complete}}}

\newcommand{\withheld}{\textcolor{trizmagenta}{\textbf{not promoted}}}
\newenvironment{tightitemize}{%
  \begin{itemize}\setlength{\itemsep}{1.5pt}\setlength{\parskip}{0pt}\setlength{\parsep}{0pt}\setlength{\topsep}{2.5pt}}
  {\end{itemize}}

\title{\triza: Guiding Agent Evolution from Pattern Recognition to Solution Invention}

\author{
    Wenyin Liu$^{1}$,
    Yiheng Huang$^{2}$,
    Kai Wang$^{3}$ \\[4pt]
    $^{1}$ Guangdong University of Technology, \texttt{liuwy@gdut.edu.cn} \\
    $^{2}$ Beijing University of Posts and Telecommunications, \texttt{huangyiheng.bupt@gmail.com} \\
    $^{3}$ Beijing Denglu Technology Ltd, \texttt{kaiwen@denglu.net.cn}
}

\begin{document}
\maketitle
\begin{abstract}
We propose \triza{} (TRIZ exponentiated by an agent), a general R\&D automation paradigm that combines TRIZ inventive theory with LLM-driven agent evolutionary search. TRIZ's 40 inventive principles and contradiction matrix provide structured, explainable directions for solution generation, replacing random or untyped mutation with theory-guided ideation. Functional information (FI), operationalized under a frozen reference contract, is combined with TRIZ Ideality to measure useful and harmful function on a commensurable information scale, while hard gates keep promotion distinct from metric improvement. We validate \triza{} in cybersecurity, where the threat landscape is adversarial and rapidly evolving, on three well-known datasets: PowerDuck GOOSE, CICIoT2023, and CIC-DDoS2019. Under paired-rerun protocols with protocol fingerprinting and hard-gate validation, the legacy experiments yield absolute F1 improvements of $+2.88$, $+4.23$, and $+0.15$ percentage points, respectively. A completed 45-activity CICIoT2023 campaign further increases macro-F1 from $0.8325$ to $0.8483$, but does not pass its frozen promotion gate. Every result remains traceable from contradiction identification and TRIZ principle selection to code transformation, evaluation metrics, and promotion decision.
\end{abstract}

\section{Introduction}

AI-driven discovery is beginning to move beyond pattern recognition. AlphaFold transformed protein-structure prediction \citep{jumper2021}; FunSearch coupled language-model proposals to program evaluation and discovered new mathematical constructions \citep{romeraparedes2024}; AlphaEvolve extended the same generate-and-test paradigm to algorithms and systems \citep{novikov2025}. These systems demonstrate that an LLM can participate in an evolutionary loop: propose an artifact, execute it, score it, and use the result to generate another proposal.

Yet this loop is still largely directed by evaluation after the fact. A score can rank two candidates, but it does not identify the contradiction that blocks progress, explain why a particular transformation is appropriate, or distinguish a genuine technical advance from a candidate that learned to exploit its evaluator. The missing capability is not generation but a theory of directed invention: a mechanism that states \emph{where to search}, a measurement that states \emph{what kind of progress occurred}, and a traceable record that allows an independent reviewer to trace the evidence and rules behind each promotion decision.

This gap is especially visible in cybersecurity R\&D. Detection quality must be improved without unacceptable false positives, missed attacks, leakage, runtime cost, or split contamination. The task is therefore not simply to maximize F1 but to resolve this technical contradiction under a fixed evaluation protocol that evaluates both the intended improvement and its associated adverse effects while enforcing the relevant constraints. Optimizing a single scalar metric can yield a candidate that scores highly, but should not displace the incumbent (current best solution).

\triza{} addresses this problem by combining two complementary ideas. First, TRIZ organizes invention around technical contradictions, 40 recurring inventive principles, and Ideality derived from the Ideal Final Result (IFR) \citep{altshuller1984}. We convert the principles from verbal prompts into typed code-transformation operators. Each candidate must declare a contradiction and an operator such as Segmentation, Local Quality, Prior Action, or Parameter Change before its result is known. Second, functional information (FI) quantifies, relative to a predefined reference population, how rare it is for a configuration to achieve at least a specified level of function \citep{hazen2007}. The Law of Increasing Functional Information (LIFI) emphasizes the joint roles of configuration, selection, and function in evolving systems \citep{wong2023}. To apply FI to engineering search, we define the reference population, qualification criteria, adverse-effect metrics, and evaluation procedure before the search begins and keep them fixed throughout the search.

Inventive search needs both a proposer and a judge. TRIZ supplies structured directions for producing candidates; FI/Ideality supplies a common information scale for describing useful and harmful achievement. Neither is allowed to erase hard constraints. Integrity, reproducibility, safety, and resource gates are checked before promotion, and negative decisions are retained as evidence. This separation is central to the paper: an improved metric is an observation, whereas promotion is a governed decision.

Our contributions are:
\begin{tightitemize}
    \item \textbf{Theory-guided proposal operators.} We map TRIZ principles to typed transformation contracts that specify which transformations an LLM may apply and what evidence is required to evaluate the resulting candidate.
    \item \textbf{Operational FI/Ideality.} Useful and harmful functions are represented on a common bit scale under a named protocol; FI is used to measure functional achievement rather than as a standalone objective to be maximized independently of utility and constraints.
    \item \textbf{Evidence-bound promotion.} Hard-gate checks are applied before beneficial and adverse effects are compared, ensuring that a higher aggregate score cannot compensate for data leakage, unsafe operation, or incomplete replication.
    \item \textbf{End-to-end traceability.} Each candidate has an audit record containing the identified contradiction, selected TRIZ principle, applied typed transformation, evaluation-protocol fingerprint, metric values, and controller decision.
    \item \textbf{Layered cybersecurity evaluation.} We report four evidence layers separately: legacy paired results, a completed CICIoT2023 campaign, a controller-policy audit, and negative results from subsequent optimization. These layers answer different scientific questions and therefore support different conclusions.
\end{tightitemize}

\vspace{-5mm}
\section{Related Work}

\textbf{TRIZ and computational invention.} TRIZ systematizes invention through contradictions, engineering parameters, Ideality, and reusable principles \citep{altshuller1984}; later work adapts these ideas to software and service systems \citep{mann2004}. Recent LLM systems automate parts of the TRIZ reasoning process or coordinate TRIZ-specialized agents \citep{jiang2025,szczepanik2025}. These systems mainly use TRIZ to structure the ideation process. \triza{} makes a narrower and testable claim: the selected principle becomes the declared type of a mutation inside an executable code-evolution loop, and that declaration is recorded before evaluation.

\textbf{LLM-guided program evolution.} FunSearch and AlphaEvolve show that LLM-generated program variants can be executed, evaluated, and selected to improve program performance \citep{romeraparedes2024,novikov2025}. OmniScientist extends automation across the research lifecycle and human--agent collaboration; AutoSOTA links research papers to their code repositories and couples reproduction, repair, ideation, evaluation, and supervised improvement \citep{shao2025omniscientist,li2026autosota}. These systems automate a broad range of research activities, whereas \triza{} focuses on how candidate transformations are specified and constrained. Rather than asking LLM only to \emph{generate a promising program variant}, it asks LLM to \emph{instantiate a transformation applicable to the identified contradiction}, bounded by the selected TRIZ type, frozen protocol, and evidence obligations.

\textbf{Function-conditioned selection.} Functional information is the negative logarithm of the fraction of configurations that achieve a specified level of function \citep{hazen2007}. It is defined relative to a predefined reference population and a qualification threshold. Wong et al.\ emphasize how variation and selection can increase functional information across evolving systems \citep{wong2023}. However, this does not imply that FI increases monotonically across candidates or that rarity alone makes a candidate useful. In our current experiments, we specify the qualification criteria, reference population, sample size, and stopping rule before evaluation. Confidence sequences that remain statistically valid across repeated interim analyses may support future experiments with data-dependent stopping \citep{howard2021}.

\section{Methodology}

\subsection{Framework Overview}

\begin{figure}[!t]
  \centering
  \includegraphics[width=\linewidth]{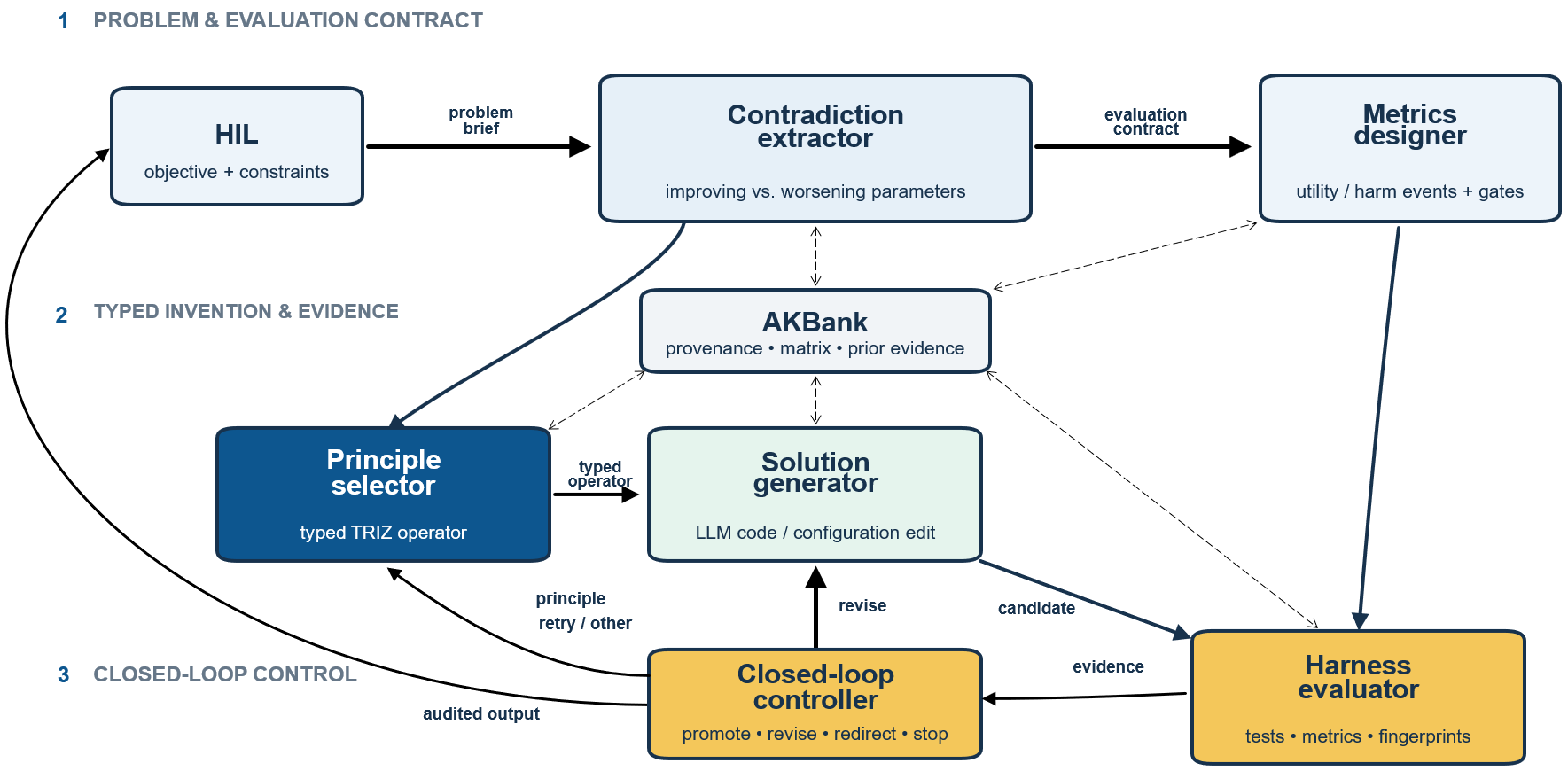}
  \caption{System architecture of \triza{}.}
  \label{fig:architecture}
\end{figure}

Figure~\ref{fig:architecture} shows the system as a closed invention loop. It contains four stages organized in three functional layers: (1) Problem formulation and evaluation contract (Stage 1 in light blue color) specifying the problem, constraints, and evaluation criteria; (2) Typed invention and evidence (Stage 2 and Stage 3, deep blue and light green, respectively) performing main inventive operations as the core of the system; (3) Closed-loop control (Stage 4, yellow) evaluating candidate solutions and feeding results back to guide the next iterations. \textbf{Stage 1: contradiction identification.} A human-in-the-loop (HIL) brief specifies the target function and non-negotiable constraints. The contradiction extractor analyzes the brief and identifies a technical contradiction in terms of the parameter to be improved and the parameter that may worsen as a consequence, and testable criteria for the intended benefit and the associated adverse effect. The metrics designer encodes these criteria in an executable evaluation protocol and freezes the protocol before candidate generation. \textbf{Stage 2: principle selection.} The Authority Provenance Knowledge Bank (AKBank) supplies source-tagged TRIZ mappings, prior outcomes, and measurement contracts. The selector retrieves the TRIZ principles recommended for the identified contradiction. \textbf{Stage 3: solution generation.} An LLM instantiates one selected principle as a code or configuration change. \textbf{Stage 4: evaluation and evolution.} The harness executes the candidate, after which the controller either promotes or revises it, redirects the principle search, or escalates the candidate and its audit record to an HIL operator for review and decision; any proposed high-risk action requires explicit authorization from a human operator with the required authority before execution.

The architecture deliberately separates knowledge, generation, and decision. AKBank can recommend an operator, but it cannot certify the resulting candidate. The LLM can write an edit, but it cannot choose a favorable metric after observing the output. The controller can compare evidence, but it cannot rewrite the frozen contradiction or protocol. These boundaries turn a persuasive agent conversation into a reviewable engineering process.

\subsection{Theory-Guided Ideation}

\triza{} differs from untyped program evolution in how it generates
candidate transformations: each transformation must instantiate a
selected TRIZ principle and conform to the corresponding typed contract. Given a parent artifact $s$, an identified technical contradiction $c$, a selected principle $r$, and a frozen evaluation protocol $\kappa$, we define the typed proposal record as
\begin{equation}
    \pi=(id(s),c,r,\tau_r,\Delta,\kappa),
    \qquad
    s'=\tau_r(s;\Delta),
    \label{eq:proposal}
\end{equation}
where $\tau_r$ is the typed transformation operator associated with
principle $r$, and $\Delta$ contains its concrete arguments. The
instantiated transformation $\tau_r(\cdot;\Delta)$ maps the parent
artifact $s$ to the candidate artifact $s'$. The operator specifies its input and output types, what the transformation may not change, and the evidence required for evaluation. The LLM selects concrete arguments and generates the corresponding code edit, but the execution record must show that the resulting diff conforms to the declared operator. A transformation is considered TRIZ-guided only if its operator type is declared before evaluation; a favorable result alone is insufficient.

Table~\ref{tab:mapping} gives representative mappings. They are broader than individual hyperparameters but narrower than open-ended prompts. For example, Local Quality authorizes context-specific behavior under a registered context type instead of arbitrary architecture search. The complete 40-principle map appears in Appendix~B.

\begin{table}[t]
\caption{Representative TRIZ principles as executable transformation contracts.}
\label{tab:mapping}
\centering
\small
\renewcommand{\arraystretch}{1.16}
\begin{tabularx}{\linewidth}{>{\raggedright\arraybackslash\bfseries}p{0.90in}>{\raggedright\arraybackslash}p{2.00in}>{\raggedright\arraybackslash}X}
\toprule
\rowcolor{trizblue!15}
\textcolor{triznavy}{Principle} & \textcolor{triznavy}{Typed transformation} & \textcolor{triznavy}{Cybersecurity example} \\
\midrule
Segmentation (1) & Partition components or decisions behind a stable interface. & Per-family detectors with a shared output schema. \\
Local Quality (3) & Dispatch context-specific behavior under a registered context. & Class-aware scaling or decision boundaries. \\
\rowcolor{trizrow}
Prior Action (10) & Prepare reusable state bef critical path. & Train-only feature cache bound to a split digest. \\
Parameter Change (35) & Change a type, threshold, loss, weight, or operating point. & Balanced class weights or calibrated loss. \\
\bottomrule
\end{tabularx}
\end{table}

TRIZ does not guarantee that a selected operator will work. Its value is directional: rather than asking the LLM to generate another unconstrained variant, TRIZ specifies a transformation derived from an established principle and tests whether the resulting candidate resolves the stated contradiction. Failed proposals remain informative because they rule out an operator instantiation under a known contract and help identify which constraint or adverse effect should be addressed in the next iteration.

\subsection{Functional Information and FI/Ideality}

TRIZ specifies the transformations used to generate candidates, whereas FI compares the useful and harmful performance of each resulting candidate with a fixed reference population. An evidence level $l$ specifies a tier of validation requirements under the frozen protocol $\kappa$. These requirements may include the prescribed data split, number of repeated runs,
independent replication, and Boolean qualification tests. Let
$\mathcal{B}_l$ denote the fixed reference population evaluated according to the requirements of evidence level $l$, with
$N_l=|\mathcal{B}_l|$. For a candidate $s$ and a reference configuration $x\in\mathcal{B}_l$, write $x\succeq_U s$ if $x$ matches or exceeds the useful-function performance attained by $s$ according to the qualification criteria specified for level $l$. We estimate useful-function information as
\begin{equation}
I_{U,l}(s)=-\log_2\!\left[
\frac{
1+\sum_{x\in\mathcal{B}_l}\mathbf{1}[x\succeq_U s]
}{
1+N_l
}
\right].
\label{eq:usefulfi}
\end{equation}

A large $I_{U,l}(s)$ means that few configurations in the reference
population achieve the useful-function level attained by $s$. Here, $N_l$
counts the configurations actually included in $\mathcal{B}_l$, rather than
all theoretical combinations of TRIZ principles and parameter values. Each
configuration in $\mathcal{B}_l$ must be generated by the frozen
transformation grammar and evaluated under protocol $\kappa$. The add-one
correction keeps the point estimate finite when no reference configuration
matches or exceeds $s$. For conservative comparisons, we use
$I_{U,l}^{\mathrm{LB}}(s)=-\log_2 p_{U,l}^{\mathrm{UB}}(s)$, where
$p_{U,l}^{\mathrm{UB}}(s)$ is an upper confidence bound on the proportion of
reference configurations that match or exceed $s$.

Harm is evaluated separately. Write $x\succeq_H s$ if reference
configuration $x$ is at least as harmful as $s$ under the harm comparison
specified for level $l$. The comparison uses prespecified measures such as
detector false-positive rate, attack-recall loss, data leakage, and
resource-limit violations. We define harmful-function information as
\begin{equation}
I_{H,l}(s)=-\log_2\!\left[
\frac{
1+\sum_{x\in\mathcal{B}_l}\mathbf{1}[x\succeq_H s]
}{
1+N_l
}
\right].
\label{eq:harmfi}
\end{equation}

A candidate with low measured harm has many reference configurations that
are at least as harmful and therefore has a small $I_{H,l}(s)$. By contrast,
a candidate with unusually high harm is matched or exceeded by few reference
configurations and therefore has a large $I_{H,l}(s)$. Useful and harmful
performance are both measured in bits but reported separately, so an
improvement in useful performance cannot numerically offset an increase in
harm.

An evidence level specifies the validation requirements to be applied,
whereas a candidate's certification level records the highest evidence level
whose requirements the candidate has passed. We define
\begin{equation}
L_{\mathrm{cert}}(s)
=
\max\left\{
l \,\middle|\,
s \text{ passes all validation requirements and hard gates at level } l
\right\}.
\label{eq:certlevel}
\end{equation}
We set $L_{\mathrm{cert}}(s)=-1$ when $s$ has not been certified at any
level. Thus, a candidate may be evaluated against the requirements of evidence level \(l\) before it is certified at that level.

For diagnosis and tie-breaking among candidates with the same certification
level and replication status, \triza{} computes an FI-based version of TRIZ
Ideality at the relevant evidence level:
\begin{equation}
J_l(s)=
\frac{
I_{U,l}^{\mathrm{LB}}(s)
}{
\max\left\{
\epsilon,\,
I_{H,l}^{\mathrm{UB}}(s)+\lambda_C C_{\mathrm{path}}(s)
\right\}
}.
\label{eq:ideality}
\end{equation}
Here, $\epsilon>0$ prevents division by zero. All reported runs set
$\lambda_C C_{\mathrm{path}}(s)=0$, so artifact cost does not affect the
reported comparisons.

The value of $J_l$ neither determines certification nor overrides the
mandatory checks. The controller first applies the data-integrity, safety,
reproducibility, and resource checks. A candidate that fails any of these
checks cannot be promoted. The controller then considers certification level
and replication status. Useful FI, harmful FI, cost, and Ideality are used
only to compare candidates with the same certification level and replication
status.

FI is therefore not used to reward rarity alone. A random or overfitted
artifact may be rare but still fail the prespecified qualification criteria.
These criteria specify the performance thresholds that a candidate must meet to qualify as useful. Reporting
useful and harmful FI separately prevents an improvement in useful
performance from masking an increase in harm. The Campaign08 audit in
Section~4.3 illustrates this distinction: a universal safety-oriented veto
suppressed useful recall, whereas the balanced controller evaluated recall
and the prespecified harm measures separately and achieved a better measured
trade-off between them.

\subsection{Traceability and Evidence State}
Every execution creates a claim-linked record containing parent and candidate identifiers, contradiction and principle IDs, typed operator arguments, normalized diff, protocol fingerprint, raw metric references, FI events and bounds, gate outcomes, and final decision. Human-readable strategy and append-only execution evidence are stored separately. Appendix~A defines the public registry for \texttt{project.toml}, \texttt{triz\_context.json}, \texttt{triz\_strategy\_plan.md}, and \texttt{scores.jsonl}.

The controller evolves an evidence-bearing artifact, not code alone. Level 0 covers static checks and tests; Level 1 is a paired run on a frozen split; Level 2 adds registered seeds, families, or scenarios; Level 3 adds an independently locked rerun or external evaluation. A source diff without a protocol and evidence references cannot be promoted, even if it looks promising. A rejected candidate remains in the archive with the reason for rejection. This allows the next proposal to address the identified bottleneck rather than obtain a more favorable result by changing the evaluation protocol.

Traceability establishes inspectability, not component-level causality. In the PowerDuck case, the record connects a coverage-versus-few-shot contradiction to Local Quality (3) and Parameter Change (35), a train-only \texttt{StandardScaler + LinearSVC} instantiation, and weighted-F1 $0.9694\rightarrow0.9982$. This chain shows what was proposed and executed. It does not prove that either principle alone caused the gain without an equal-budget unguided control. Appendix~E states the complete reconstruction and this boundary.

\section{Experiments}

\subsection{Experimental Setup and Paired Protocol}

We evaluate \triza{} on three cybersecurity tasks with different contradictions. \textbf{PowerDuck GOOSE} contains attacks in substation network traffic \citep{zemanek2022}; the task is few-shot classification across 15 registered classes and is scored by weighted-F1. \textbf{CICIoT2023} contains large-scale IoT attacks \citep{neto2023}; we evaluate held-out attack families with macro-F1, attack recall, and detector FPR. \textbf{CIC-DDoS2019} contains multiple DDoS families \citep{sharafaldin2019}; its final registered evaluation reports weighted-F1 over all ten folds.

All reported candidate–baseline differences were obtained by rerunning both under the same evaluation settings. A hash records the dataset version, data split, preprocessing, training-only feature fitting, metric implementation, random seed, and candidate budget. Results are treated as a paired comparison only when their hashes match. Before a score can influence promotion, the harness checks tests, fingerprint identity, frozen representation budget, zero train-test overlap, and campaign-specific safety gates. Detector FPR measures how often the detector incorrectly labels benign samples as attacks, whereas promotion specificity measures how often the controller correctly rejects candidates that should not be promoted. They describe different decision levels and are not interchangeable.

\subsection{Legacy Paired Outcomes}

Table~\ref{tab:evidence} reports an evidence ladder rather than a single leaderboard. PowerDuck is a complete paired observation but its near-ceiling result should not be read as proof that the task is solved in deployment. The legacy CICIoT value is a later downstream incumbent, not the formal V2 promotion decision. CIC-DDoS is a complete ten-fold paired evaluation covering its full registered scope.

\begin{table}[t]
\caption{Evidence ladder. Results with different decision boundaries are not merged.}
\label{tab:evidence}
\centering
\small
\renewcommand{\arraystretch}{1.14}
\begin{tabularx}{\linewidth}{>{\raggedright\arraybackslash\bfseries}p{1.00in}>{\raggedright\arraybackslash}p{1.30in}>{\raggedright\arraybackslash}p{1.48in}>{\raggedright\arraybackslash}X}
\toprule
\rowcolor{trizblue!15}
\textcolor{triznavy}{Task / layer} & \textcolor{triznavy}{Comparison} & \textcolor{triznavy}{Observed evidence} & \textcolor{triznavy}{Status and boundary} \\
\midrule
PowerDuck legacy & Paired weighted-F1 & $0.9694\rightarrow0.9982$ ($+2.88$ pp) & \complete; no principle-level attribution \\
\rowcolor{trizrow}
CICIoT legacy & Paired macro-F1 & $0.8326\rightarrow0.8749$ ($+4.23$ pp) & downstream incumbent, not formal V2 \\
CIC-DDoS legacy & Paired weighted-F1 & $0.7470\rightarrow0.7485$ ($+0.15$ pp) & \complete; final ten-fold evaluation \\
\rowcolor{trizrow}
CICIoT formal V2 & 45-activity campaign & macro-F1 $+0.0158$; recall $+0.018$; FPR $-0.0017$ & complete evidence; \withheld \\
Campaign08 audit & 2,040 controller decisions & balanced policy: promotion F1 $0.811$, specificity $0.989$ & descriptive nested aggregate \\
\rowcolor{trizrow}
V17/V18 downstream & Multi-metric sequence & V17 retained for study; V18 deteriorated on three metrics & negative evidence preserved \\
\bottomrule
\end{tabularx}
\end{table}

The three legacy results show that \triza{} produced candidates with higher scores than their rerun baselines on all three tasks when each candidate and baseline were evaluated under the same protocol. However, the comparisons do not determine whether the improvements came from TRIZ guidance, FI/Ideality, or another part of the workflow. The large PowerDuck and CICIoT changes show that the workflow can find useful candidates where class coverage and distribution shift create room for structured transformations. The complete ten-fold CIC-DDoS result is equally important: It shows that \triza{} does not convert every task into a dramatic gain and that the evaluation preserves its preregistered scope even when the observed improvement is small.
\subsection{Completed CICIoT2023 V2 and FI/Ideality Promotion Audit}

The formal CICIoT2023 V2 campaign completed all 45 registered activities on a frozen candidate. FI/Ideality-guided selection identified a configuration that increased macro-F1 from $0.8325$ to $0.8483$ (absolute $+0.0158$, $1.90\%$ relative). Attack recall increased by $0.018$, detector FPR decreased by $0.0017$, 43 of 45 family-seed groups improved, and all five seed blocks were positive. Together, the higher F1 and recall and the lower FPR indicate improvements in both benefits and harms, rather than in F1 alone.

Nevertheless, the candidate failed two frozen promotion thresholds: the relative macro-F1 gain was below $3.2277\%$, and the detector-FPR reduction was below $0.005$. The controller therefore withheld promotion. This is not a contradiction in the method. FI/Ideality shapes which useful and harmful evidence is valued; the hard gate decides whether that evidence is sufficient to replace the incumbent.

\begin{figure}[!t]
  \centering
  \includegraphics[width=\linewidth]{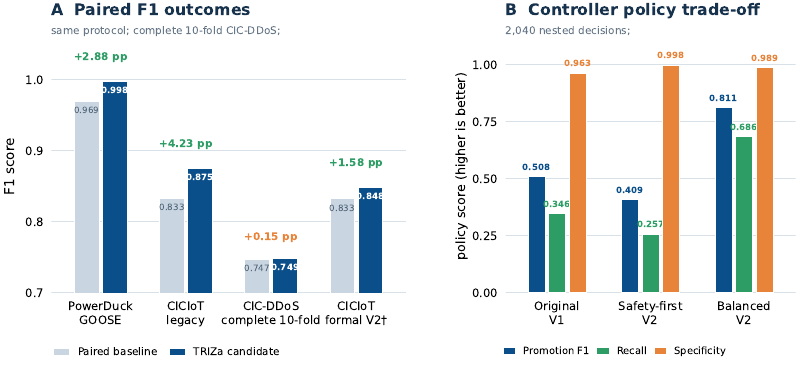}
  \caption{Experimental evidence at two decision levels. (A) Final same-protocol F1 outcomes, including the complete ten-fold CIC-DDoS evaluation and formal V2; the formal V2 candidate was not promoted. (B) Campaign08 policy trade-off. Safety-first V2 nearly maximizes specificity but loses recall and promotion F1; balanced V2 recovers useful decisions while retaining $0.989$ specificity. Campaign08 is a descriptive aggregate of 2,040 nested decisions.}
  \label{fig:results}
\end{figure}

Campaign08 tests the controller rather than the detector. Original V1 achieved promotion F1 $0.508$, recall $0.346$, and specificity $0.963$. Safety-first V2 applied a global FI/Ideality veto and raised specificity to $0.998$, but recall fell to $0.257$ and promotion F1 to $0.409$. Balanced V2 kept useful and harmful evidence as separate channels; promotion F1 rose to $0.811$, recall to $0.686$, and specificity remained $0.989$. Figure~\ref{fig:results} makes the trade-off visible. The artifact is an aggregate, so we do not treat 2,040 nested decisions as independent replicates or report ungrounded $p$-values.

\subsection{Downstream Evolution and Negative Evidence}

A later CICIoT sequence produced V17 at macro-F1 $0.8753$ and MCC $0.7755$, with detector FPR $0.0231$ but attack recall $2.43$ percentage points below the V1/RF100 reference. V18 slightly recovered recall to $0.8811$, yet macro-F1, MCC, and detector FPR all deteriorated relative to V17. The controller rejected V18. The complete four-panel sequence is reported in Appendix~D because it belongs to downstream evolution, not the formal V2 selection campaign.

Together, the four evidence layers support a precise claim. Typed TRIZ proposals can produce competitive candidates across distinct cybersecurity tasks; FI/Ideality changes which benefit-harm trade-offs the controller can see; and prespecified gates prevent an improvement in the main performance metric from overriding the required validation checks. They do not identify the causal contribution of TRIZ or FI in isolation. Traceability reconstructs the path, whereas causal attribution requires a matched-budget ablation outside the claim made here.

\subsection{What the Evidence Establishes}

The four evidence layers answer different questions. The legacy paired reruns test whether the workflow can find candidates that outperform their baselines under the same protocol; they do not identify which part of the workflow caused the improvements. Formal V2 tests one frozen candidate against the same promotion rules in all 45 registered activities. Although the candidate improved macro-F1, attack recall, and detector FPR, it failed two promotion thresholds and was not promoted. Campaign08 compares controller policies, whereas the V17/V18 sequence shows how the controller handles a later candidate that improves one measure but worsens others.

Before promotion, the controller checks that the dataset, split, preprocessing, code, seed, and resource records match the registered protocol. It then compares measured benefits and harms and applies the campaign-specific gates. A higher macro-F1 therefore cannot compensate for data leakage, a failed safety check, or missing replication.

These experiments show that the workflow can be applied across three cybersecurity tasks and can record why each candidate was retained or rejected. They do not determine how much TRIZ, FI/Ideality, or the controller contributed to the observed gains because the study does not include matched-budget ablations of these parts. Candidates that improve some measures but fail a gate remain in the record, allowing later proposals to address the failed requirement without changing the evaluation rules.

\subsection{Traceability and Reproducibility}

For each candidate, the recorded evidence links the diagnosed contradiction, selected TRIZ principle, applied transformation, evaluation protocol, measured results, gate outcomes, and controller decision. These records allow the reported decision to be checked against the measured results. Appendix~\ref{app:reproducibility} describes the files, evidence boundaries, and limits of reproduction.
\section{Conclusion}

\triza{} makes invention theory executable inside an agent loop: TRIZ supplies typed search directions, FI/Ideality describes useful and harmful achievement under a frozen contract, and hard gates decide promotion. The cybersecurity studies show paired rerun gains, a completed CICIoT2023 candidate that improved three detector measures but was correctly not promoted, and a controller audit in which promotion balanced evidence on benefits and harms recovered  quality without abandoning promotion specificity. Results from candidates that are not promoted, as well as results that apply only under specified conditions, are also retained as complete audit records.

Our main contribution is an auditable workflow that records how each candidate is guided, generated, evaluated, and selected. Neither TRIZ nor FI alone causes every performance gain. For each reported outcome, the audit trail links the identified contradiction, selected TRIZ principle, applied transformation, evaluation protocol, measured results, gate outcomes, and controller decision. \triza{} thus replaces unconstrained trial-and-error over metrics with a reviewable process in which TRIZ principles guide candidate transformations and the evidence for each decision is preserved.

\noindent\textbf{AI system disclosure.} The human authors originated the goal of using AI to operationalize TRIZ for invention and provided the evolving research directions. Human-initiated
extensions included FI-based formulation of Ideality, design of the loop controller, and mechanisms for tracing candidate transformations and decisions. AI coding agents implemented the system, ran computational experiments, produced figures, and drafted the manuscript. The human authors reviewed the outputs, extensively revised the manuscript and take responsibility for its final content.

\clearpage
\bibliography{triza_refs}
\bibliographystyle{icais_conference}

\clearpage
\appendix
\begingroup
\footnotesize
\raggedbottom
\setlength{\parskip}{0pt}
\setlength{\textfloatsep}{4pt plus 1pt minus 1pt}
\setlength{\floatsep}{4pt plus 1pt minus 1pt}
\setlength{\intextsep}{4pt plus 1pt minus 1pt}
\setlength{\abovedisplayskip}{3pt plus 1pt minus 1pt}
\setlength{\belowdisplayskip}{3pt plus 1pt minus 1pt}
\setlength{\abovecaptionskip}{2pt}
\setlength{\belowcaptionskip}{0pt}
\setlength{\tabcolsep}{2.4pt}
\setlength{\LTpre}{2pt}
\setlength{\LTpost}{2pt}
\renewcommand{\arraystretch}{0.98}
\titlespacing*{\section}{0pt}{6pt plus 1pt minus 1pt}{2pt}
\titlespacing*{\subsection}{0pt}{4pt plus 1pt minus 1pt}{1.5pt}

\section{Traceability Artifacts}
\label{app:traceability}

\subsection{Traceability Record Schema}

\triza{} separates human-readable plans from append-only evaluation
records. Table~\ref{tab:registry} defines the filenames, fields, and roles
used by the traceability schema. These files are not included in the
current submission package but will be released in a public artifact
package upon publication. The released \texttt{SHA256SUMS.txt} file will
provide a byte digest for each included file. Claims in this paper remain
limited to the aggregate evidence reported in the manuscript.

\begin{table}[!htbp]
\caption{Traceability record schema and planned archive-relative filenames.}
\label{tab:registry}
\centering
\scriptsize
\renewcommand{\arraystretch}{0.98}
\begin{tabular}{
    >{\raggedright\arraybackslash}p{1.05in}
    >{\raggedright\arraybackslash}p{2.35in}
    >{\raggedright\arraybackslash}p{1.55in}}
\toprule
\textbf{Artifact} & \textbf{Recorded content} & \textbf{Role} \\
\midrule

\path{project.toml}
&
claim ID, task, constraints, data and split IDs, protocol ID, and budgets
&
project and protocol record
\\

\path{triz_context.json}
&
contradiction, improving and worsening parameters, principle set, and sources
&
contradiction record
\\

\path{triz_strategy_plan.md}
&
declared transformations, arguments, prohibited changes, and required evidence
&
typed transformation plan
\\

\path{scores.jsonl}
&
parent and candidate IDs, protocol ID, transformations, seeds, metrics,
FI bounds, evidence levels, gates, and decisions
&
append-only evaluation record
\\

\path{SHA256SUMS.txt}
&
byte digests for files included in the released artifact package
&
file identity and integrity check
\\

\bottomrule
\end{tabular}
\end{table}

\subsection{Canonical Paper-Level Record}

The following paper-level record summarizes only the PowerDuck values and
claim boundaries reported in this manuscript. It is not a raw execution
record and therefore does not include run identifiers, raw predictions,
or file hashes.

\begin{scriptsize}
\begin{verbatim}
{
  "claim_id": "powerduck.weighted_f1.paired",
  "task": "PowerDuck GOOSE",
  "comparison": "same_protocol_paired_rerun",
  "contradiction_id": "coverage_vs_few_shot",
  "principles": [3, 35],
  "operator": "local_quality.parameter_change",
  "metrics": {
    "baseline_weighted_f1": 0.9694,
    "candidate_weighted_f1": 0.9982
  },
  "scope": {
    "registered_classes": 15
  },
  "gates": {
    "protocol_match": true,
    "integrity": true
  },
  "decision": "retain_complete_paired_observation",
  "boundary": "no_principle_level_causal_claim"
}
\end{verbatim}
\end{scriptsize}

\section{Full TRIZ-to-Transformation Mapping}

The mapping below defines operator classes, not guaranteed improvements. Each concrete operator must further specify input/output type, admissible arguments, prohibited changes, and evidence obligations.

\small
\begin{longtable}{p{0.35in}p{1.45in}p{3.35in}}
\caption{TRIZ principles and software/agent transformation classes.}\label{tab:fullmap}\\
\toprule
\textbf{No.} & \textbf{Principle} & \textbf{Typed transformation class} \\
\midrule
\endfirsthead
\toprule
\textbf{No.} & \textbf{Principle} & \textbf{Typed transformation class} \\
\midrule
\endhead
1 & Segmentation & Partition modules, data, decisions, or search regions behind a stable interface. \\
2 & Taking out & Isolate a confounder, harmful path, privileged component, or expensive stage. \\
3 & Local quality & Dispatch context-specific policy under a registered context type. \\
4 & Asymmetry & Apply different rules to unequal classes, risks, or resource directions. \\
5 & Merging & Compose compatible stages, features, models, or evaluation passes. \\
6 & Universality & Replace duplicated components with one multi-role typed service. \\
7 & Nested doll & Introduce nested scopes, hierarchical models, or encapsulated subpipelines. \\
8 & Anti-weight & Compensate load, bias, latency, or resource pressure with a countermeasure. \\
9 & Preliminary anti-action & Install a guard, rollback point, adversarial check, or preventive constraint. \\
10 & Prior action & Precompute, cache, initialize, or validate a state before the critical path. \\
11 & Beforehand cushioning & Add fallback, buffer, retry budget, or damage-limiting boundary. \\
12 & Equipotentiality & Normalize representations or equalize operating conditions. \\
13 & The other way round & Invert control, data flow, labels, optimization direction, or dependency. \\
14 & Spheroidality/curvature & Replace brittle discrete geometry with smooth or continuous structure. \\
15 & Dynamics & Adapt a parameter or module at runtime within frozen bounds. \\
16 & Partial/excessive action & Use approximation, overprovisioning, or a bounded partial result. \\
17 & Another dimension & Add a feature, hierarchy, temporal axis, view, or latent coordinate. \\
18 & Mechanical vibration & Introduce controlled probing, perturbation, or oscillatory scheduling. \\
19 & Periodic action & Re-evaluate, refresh, checkpoint, or sample on a registered schedule. \\
20 & Continuity of useful action & Stream or persist useful work while avoiding idle transitions. \\
21 & Skipping & Add an early exit, fast path, pruning rule, or bypass for safe cases. \\
22 & Blessing in disguise & Convert errors, attacks, or rejected candidates into training or test evidence. \\
23 & Feedback & Update a bounded controller from measured residuals or evaluator evidence. \\
24 & Intermediary & Insert an adapter, proxy, translator, broker, or validation layer. \\
25 & Self-service & Add self-monitoring, self-configuration, or automated maintenance. \\
26 & Copying & Use a surrogate, mock, digital twin, replay, or immutable snapshot. \\
27 & Cheap short-lived objects & Use disposable sandboxes, ephemeral workers, or low-cost probes. \\
28 & Mechanics substitution & Replace a mechanism, representation, solver, or execution substrate. \\
29 & Pneumatics/hydraulics & Introduce elastic resource pools, queues, or flow-control abstractions. \\
30 & Flexible shells/thin films & Add a lightweight interface, wrapper, boundary, or soft constraint. \\
31 & Porous materials & Use sparse structure, slots, sampling windows, or selective permeability. \\
32 & Color changes & Change encoding, annotation, visibility, or diagnostic representation. \\
33 & Homogeneity & Align schemas, units, interfaces, distributions, or component types. \\
34 & Discarding/recovering & Garbage-collect, reset, recycle, or restore a component after use. \\
35 & Parameter changes & Change a type, range, threshold, loss, weight, or operating point. \\
36 & Phase transitions & Switch between discrete modes or representations at a registered boundary. \\
37 & Thermal expansion & Scale capacity or spacing as load changes within resource gates. \\
38 & Strong oxidants & Increase validation strength, selection pressure, or stress testing. \\
39 & Inert atmosphere & Isolate execution in a sandbox, no-op mode, or low-interaction context. \\
40 & Composite materials & Construct an ensemble or layered architecture with typed interfaces. \\
\bottomrule
\end{longtable}
\normalsize

\section{Evaluation Contracts and Experimental Details}

\subsection{Frozen Contract}

A valid comparison binds dataset version and license; inclusion and exclusion rules; split indices; preprocessing and feature-fit scope; candidate budget; seeds; primary and secondary metrics; useful and harmful qualification events; resource ceilings; missing-data behavior; stopping rule; and code and environment digests. The protocol fingerprint is the SHA-256 of a canonical serialization of these fields. Changing any bound field creates a new protocol rather than silently extending the old result.

At evidence level $l$, the useful event $Q_{U,l}(s)$ and each harm event $Q_{H,l,k}(s)$ are evaluated from raw predictions. Promotion is admissible only when
\[
G_l(s)=\mathbf{1}\{g_{l,k}(s)\le 0\ \text{for every registered gate }k\}=1.
\]
No FI or Ideality value can override $G_l(s)=0$. If an adaptive process may stop after repeated inspection, an anytime-valid confidence sequence replaces a fixed-endpoint interval \citep{howard2021}.

\subsection{Dataset and Metric Contracts}

\begin{table}[!htbp]
\caption{Registered task contracts.}
\centering
\scriptsize
\begin{tabular}{>{\raggedright\arraybackslash}p{0.95in}>{\raggedright\arraybackslash}p{1.35in}>{\raggedright\arraybackslash}p{1.25in}>{\raggedright\arraybackslash}p{1.50in}}
\toprule
\textbf{Task} & \textbf{Primary metric} & \textbf{Useful event} & \textbf{Key harm/integrity checks} \\
\midrule
PowerDuck & weighted-F1, 15 classes & paired improvement under identical split & no embedding refit; class coverage; 10-shot boundary \\
CICIoT2023 & macro-F1 across held-out families & family/seed improvement and recall & detector FPR, attack recall, split and seed identity \\
CIC-DDoS2019 & weighted-F1 over folds & paired fold improvement & train-only feature fit, complete fold ledger \\
Campaign08 & promotion F1 and balanced accuracy & correct promotion decision & false promotion, nested decision dependence \\
\bottomrule
\end{tabular}
\end{table}

The formal CICIoT strict gate requires a relative macro-F1 gain of at least $3.2277\%$ and detector-FPR reduction of at least $0.005$, in addition to the integrity gates. These thresholds belong to that campaign and are not universal constants.

\section{Extended Results}

\subsection{Formal V2 and Campaign08}

\begin{table}[!htbp]
\caption{Extended CICIoT2023 evidence. Promotion metrics describe the controller, not the detector.}
\centering
\scriptsize
\begin{tabular}{p{1.6in}rrrr}
\toprule
\textbf{Configuration/policy} & \textbf{Macro-F1} & \textbf{Recall} & \textbf{Precision} & \textbf{Specificity} \\
\midrule
RF100 formal baseline & 0.8325 & -- & -- & -- \\
Formal V2 selected & 0.8483 & $+0.018$ & -- & FPR $-0.0017$ \\
Original V1 controller & -- & 0.346 & 0.952 & 0.963 \\
Safety-first V2 controller & -- & 0.257 & 0.997 & 0.998 \\
Balanced V2 controller & -- & 0.686 & 0.993 & 0.989 \\
\bottomrule
\end{tabular}
\end{table}

Formal V2 covers all 45 registered activities. Campaign08 is reported as a descriptive aggregate of 2,040 nested decisions; its decision level does not support treating the count as 2,040 independent statistical replicates.

\subsection{V14--V18 Downstream Sequence}

\begin{table}[H]
\caption{Downstream metric values shown in Figure~\ref{fig:downstream}.}
\centering
\scriptsize
\begin{tabular}{lrrrr}
\toprule
\textbf{Model} & \textbf{Macro-F1} & \textbf{MCC} & \textbf{Detector FPR} & \textbf{Attack recall} \\
\midrule
V1/RF100 & 0.8325 & 0.7153 & 0.0504 & 0.9052 \\
V14 & 0.8749 & 0.7746 & 0.0228 & 0.8779 \\
V17 & 0.8753 & 0.7755 & 0.0231 & 0.8809 \\
V18 & 0.8752 & 0.7755 & 0.0232 & 0.8811 \\
\bottomrule
\end{tabular}
\end{table}

\begin{figure}[H]
  \centering
  \includegraphics[width=.84\linewidth]{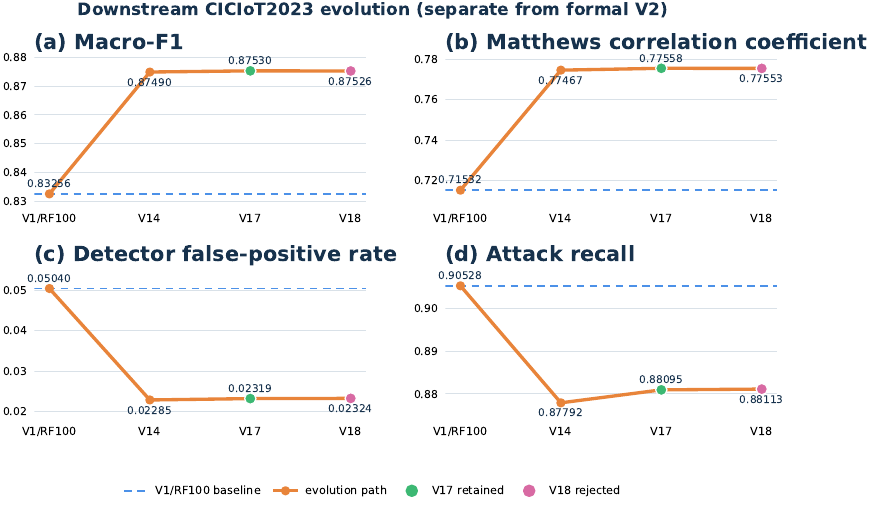}
  \caption{Downstream CICIoT2023 evolution. Relative to V1/RF100, V17 changes macro-F1 by $+5.13\%$, MCC by $+8.42\%$, and detector FPR by $-54.0\%$, while attack recall remains $2.43$ pp below baseline. V18 is rejected. This path is separate from formal V2 selection.}
  \label{fig:downstream}
\end{figure}

\section{Full Traceability Case Study}

The PowerDuck claim is organized as the following reconstruction chain:
\begin{tightitemize}
\item \textbf{Problem contract:} classify all 15 registered classes with no more than ten examples per class under a frozen embedding and split.
\item \textbf{Contradiction:} broader detection coverage is desired, but the available class-specific evidence is small.
\item \textbf{Principles:} Local Quality (3) permits class-aware behavior; Parameter Change (35) permits a registered change in scaling, loss, or class weighting.
\item \textbf{Transformation:} instantiate a train-only scaling and linear-SVM pipeline with balanced class weights.
\item \textbf{Evaluation:} rerun the baseline and candidate with identical data, split, embeddings, metric code, and seed; record weighted-F1 $0.9694\rightarrow0.9982$.
\item \textbf{Decision:} retain the candidate as a complete paired observation. Do not infer principle-level causality without an equal-budget unguided control.
\end{tightitemize}

The case study fixes the claim-level reconstruction represented in this paper. Byte-level replay is a distinct artifact property: it requires the corresponding protocol, predictions, and append-only decision record identified by the contract in Appendix~A.

\section{Reproducibility and Evidence Boundaries}
\label{app:reproducibility}

\subsection{What the Records Support}

Appendix~A defines the traceability record schema, Appendix~C specifies
the dataset and metric rules, and Appendix~E gives the PowerDuck example.
Together, these materials allow a reader to compare a reported result and
controller decision with the stated protocol and gate outcomes. This
supports claim-level checking, but it does not by itself guarantee an exact
rerun. Exact reproduction also requires access to the data or retrieval
instructions, executable code, software environment, candidate
configuration, and random seeds.

When only aggregate results are available, the manuscript identifies them
as supplied aggregates and limits its claims to the reported summaries.
Such aggregates cannot be used to recover missing decision-level data or
to treat related decisions as independent experimental repetitions.

\subsection{Evidence-Specific Boundaries}

Formal CICIoT2023 V2 completed all 45 registered activities. Macro-F1
increased from $0.8325$ to $0.8483$, attack recall increased by $0.018$,
and detector FPR decreased by $0.0017$. The candidate was not promoted
because its relative macro-F1 gain was below $3.2277\%$ and its
detector-FPR reduction was below $0.005$. The record therefore preserves
both the measured improvements and the failed promotion decision.

Campaign08 summarizes 2,040 controller decisions made within the same
experimental structure. These decisions are not 2,040 independent
experimental repetitions. We therefore report promotion F1, recall, and
specificity as descriptive aggregate results and do not calculate
$p$-values from the decision count.

The V17/V18 sequence belongs to a later optimization path and is not part
of formal V2 selection. V18 slightly increased attack recall relative to
V17 but worsened macro-F1, MCC, and detector FPR. It remains in the record
as a rejected candidate.

\endgroup

\end{document}